\PassOptionsToPackage{table}{xcolor}
\PassOptionsToPackage{numbers,compress}{natbib}

\documentclass{article}

\usepackage[main,final]{neurips_2026}

\usepackage[utf8]{inputenc}
\usepackage[T1]{fontenc}

\usepackage{graphicx}
\usepackage{xcolor} 
\usepackage{bm}
\usepackage{mathtools} 
\usepackage{amssymb}   
\usepackage{multirow}
\usepackage{booktabs}
\usepackage{float}
\usepackage{nicefrac}
\usepackage{microtype}
\usepackage{url}
\usepackage{hyperref}
\definecolor{deltagreen}{rgb}{0.0,0.5,0.0}
\definecolor{oursrow}{HTML}{E8F0FE}

\title{LocalAgent: Collaborative Agentic Verification for Fine-Grained Instance-Level Consistency}

\author{%
\parbox{0.96\textwidth}{\centering
    \textbf{
    Jiachen Guo$^{1,*}$,
    Xinshan Zhu$^{1,*}$,
    Boyun Wang$^{3,5}$,
    Yanyan Liang$^{3}$,\\
    Jun Wan$^{2,3}$,
    Feng Lin$^{4}$,
    Ajian Liu$^{2,3,4,\dagger}$
    }\\[0.55em]
    {\normalfont\normalsize\ttfamily
    * Equal Contributors, $\dagger$ Corresponding Author
    }\\[0.55em]
    {\normalfont\small
    $^{1}$ School of Electrical and Information Engineering, Tianjin University\\
    $^{2}$ MAIS, Institute of Automation, Chinese Academy of Sciences (CASIA)\\
    $^{3}$ SCSE, Macau University of Science and Technology (MUST)\\
    $^{4}$ State Key Laboratory of Blockchain and Data Security,
    Zhejiang University\\
    $^{5}$ AIX, University of Macau\\[0.25em]
    \texttt{\{gjiachen2025,xszhu\}@tju.edu.cn},
    \texttt{ajian.liu@ia.ac.cn}
    }
}}

\begin{document}

\maketitle

\begin{abstract}
Multimodal large language models (MLLMs) have made rapid progress in general visual understanding, yet they still struggle with fine-grained instance-level consistency verification, where the goal is to determine whether two images depict the exact same physical object under varying viewpoints, illumination, and backgrounds. This task requires both global semantic perception and precise examination of subtle local evidence, since semantically similar objects may differ only in small suspicious regions. To address this challenge, we propose \textbf{LocalAgent}, a collaborative agentic framework that integrates the complementary strengths of a generalist MLLM and lightweight specialist verifiers. The MLLM first reasons globally to identify suspicious local regions, which are then delegated to a Local Evidence Verifier (LEV) toolkit for fine-grained appearance and geometric matching. LEV comprises \textbf{Local Appearance Consistency (LAC)}, which estimates appearance similarity under illumination and color changes, and \textbf{Local Geometric Correspondence (LGC)}, which estimates spatial correspondence under viewpoint changes. LEV returns structured scalar evidence that the MLLM integrates for the final decision. We further introduce an auxiliary-capped reinforcement fine-tuning (RFT) strategy, in which answer correctness dominates auxiliary tool-use and format signals, enabling the policy to learn when verification is necessary. Experiments show that LocalAgent improves mean accuracy by \textbf{4.9\%} over the strongest CoT-RFT baseline while invoking a verifier on only \textbf{12.13\%} of samples and preserving the general multimodal capabilities of the underlying MLLM. The open-source repository is available at \url{https://github.com/Firefly-zoo/LocalAgent-main}.
\end{abstract}

\section{Introduction}

\begin{figure}[t]
\centering
\includegraphics[width=1\linewidth]{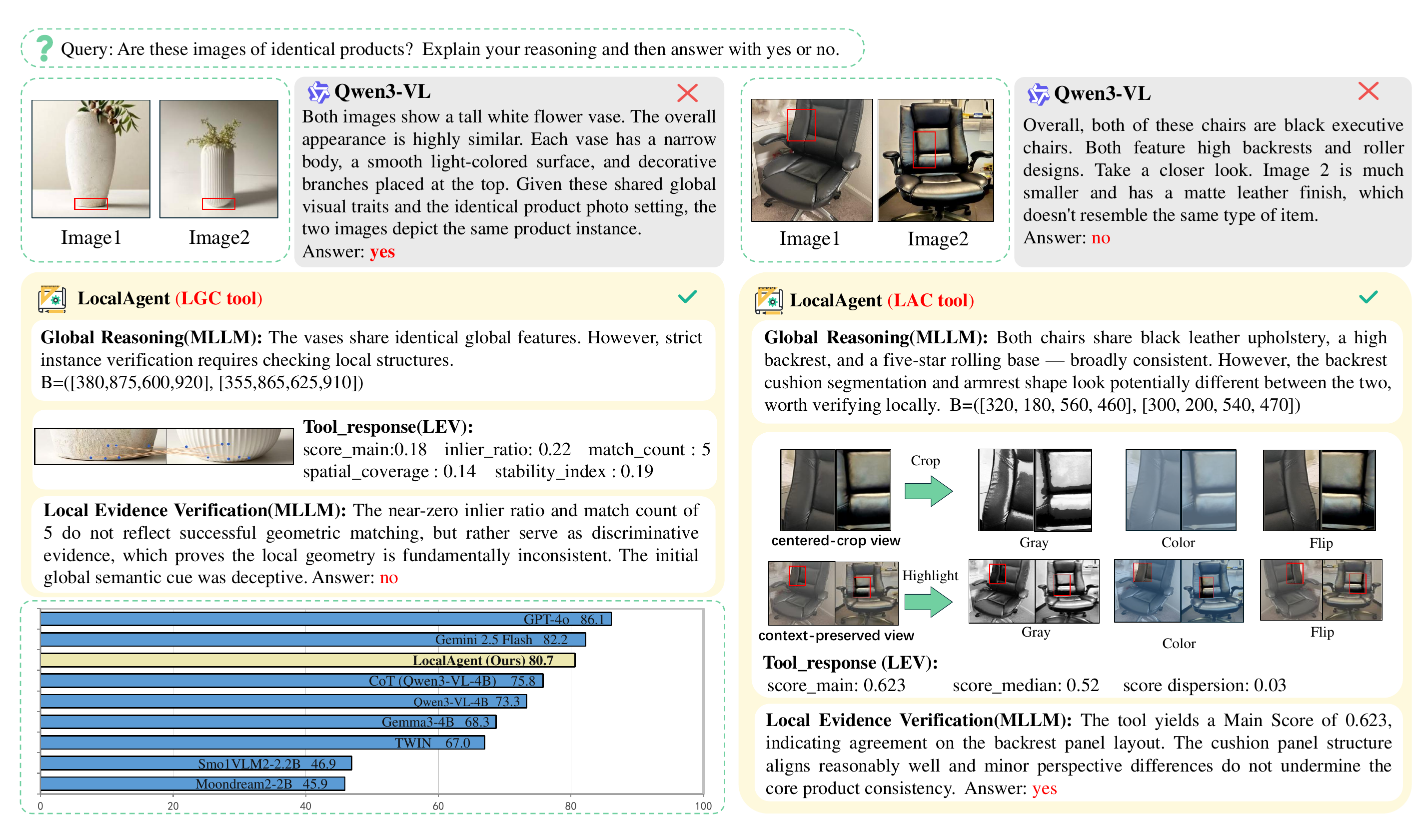}
\caption{
The motivation of LocalAgent. Standard MLLMs over-rely on global similarity and fail on instance-level verification in both directions. LocalAgent uses the MLLM as a spatial controller to localize suspicious regions, delegates them to the LEV toolkit (LAC, LGC) for invariant matching, and integrates the returned structured scalar evidence into the final decision.
}
\label{fig:motivation}
\end{figure}

Multimodal Large Language Models (MLLMs) have demonstrated remarkable proficiency in general visual understanding, yet their reliability falters in fine-grained instance-level consistency verification. This task requires the model to determine whether two images depict the exact same physical object despite drastic variations in viewpoint, illumination, and background, while a different object may look nearly identical~\cite{twin}. Unlike category-level recognition, the decision therefore often hinges on subtle local differences in texture, geometry, or brand-specific details. As illustrated in Fig.~\ref{fig:motivation}, current MLLMs can err by over-relying on coarse-grained global semantic perception while overlooking decisive local cues. Related perceptual weaknesses of MLLMs have been documented by several recent benchmarks~\cite{hallusionbench, blink, mmvp, pope, mme,wang2024journeybench}.

To address these limitations, prior research has explored chain-of-thought (CoT) reasoning, visual backtracking, and tool-augmented inference~\cite{mmreact, visuallookback, viper,dong2025insight,gao2025mmat}. While these methods enhance general reasoning robustness, they are not designed specifically for instance-level consistency verification. Moreover, providing a localized crop does not by itself guarantee reliable instance-level reasoning, because the crop is still processed by the same multimodal vision-language encoder, where subtle identity cues can remain attenuated or underutilized in the resulting semantic representation. We argue that an effective verification system should not merely increase reasoning complexity, but must strategically localize suspicious regions and delegate them to specialized modules capable of performing invariant matching beyond the MLLM's standard perceptual space.

The fundamental issue is not that MLLMs lack reasoning capacity, but that their vision encoders are optimized for cross-modal alignment. This is effective for global semantic perception, yet such encoders tend to miss fine-grained visual details~\cite{mmvp, tong2024cambrian}, such as the subtle texture, geometric, and brand-specific cues that determine instance-level identity. Motivated by this, we propose LocalAgent, which keeps the MLLM in charge of what it does well, namely global semantic perception and the localization of suspicious regions~\cite{rasheed2024glamm,ma2024groma,li2025llava}, while delegating the fine-grained discriminative judgment that lies beyond its perceptual reach to a Local Evidence Verifier (LEV) toolkit built on visual-domain features. Unlike tools that return raw visual crops, LEV provides structured, interpretable evidence that is fed back into the MLLM, enabling a closed-loop decision-making process, preventing the evidence from re-entering the same semantic bottleneck.

To train this selective verification behavior, we introduce an auxiliary-capped Reinforcement Fine-Tuning (RFT) strategy. The training objective prioritizes final answer correctness while incorporating tool invocation consistency and output format as auxiliary signals, preventing degenerate strategies such as unconditional tool invocation. We further construct weak behavioral prior labels to guide selective tool invocation without additional human annotation. Experimental results demonstrate that LocalAgent improves mean accuracy by 4.9\% over the strongest CoT-RFT baseline while invoking a verifier on only 12.13\% of samples and preserving the general multimodal capabilities of the underlying MLLM.

Our contributions are as follows.

\begin{itemize}
\item We propose LocalAgent, a collaborative framework that decomposes fine-grained instance-level consistency verification into global perception, spatial localization, and local evidence verification. The MLLM provides global semantic reasoning and identifies potentially discriminative regions, while the LEV toolkit applies appearance and geometric matching to those regions and returns structured evidence for the final decision.

\item We design structured scalar evidence through which LEV's matching results are surfaced to the MLLM as interpretable indicators rather than as visual content, allowing verification evidence to be integrated through textual reasoning without re-feeding by the vision encoder.

\item We introduce an auxiliary-capped RFT strategy with weak behavioral priors that allows the policy to learn when verification is necessary, rather than relying on hand-crafted invocation rules, improving mean accuracy by 4.9\% across six diverse data sources while invoking a verifier on only 12.13\% of samples.
\end{itemize}

\section{Related Work}
\label{sec:related}

\subsection{Fine-Grained Recognition and MLLMs}

Fine-grained visual modeling aims to distinguish instances with highly similar global appearances, where the key challenge lies in capturing subtle yet discriminative local details. Despite strong general visual understanding, MLLMs remain unreliable in fine-grained instance-level consistency verification. POPE~\cite{pope} reveals object hallucination in MLLMs, and HallusionBench~\cite{hallusionbench} exposes entangled language hallucination and visual illusion. MMVP~\cite{mmvp} and BLINK~\cite{blink} further show that MLLMs miss fine-grained visual patterns and correspondences that humans perceive easily. These weaknesses have been linked to vision encoders trained for cross-modal alignment, which can attenuate instance-specific detail~\cite{tong2024cambrian,jiang2024maven}. TWIN~\cite{twin} addresses this through targeted task formulation, yet still relies on global semantic perception without explicit mechanisms for invariant matching under complex physical transformations. LocalAgent reframes this as a collaborative verification process, leveraging the MLLM as a spatial controller to localize suspicious regions while delegating precise invariant matching to specialized verifiers trained on visual signals alone, with results surfaced as structured scalar evidence.

\subsection{Multimodal Reasoning and Tool-Augmented Inference}

MLLMs have been augmented with chain-of-thought reasoning~\cite{mmreact}, visual backtracking~\cite{visuallookback}, and tool invocation~\cite{viper} to enhance decision-making. Frameworks such as MM-REACT~\cite{mmreact} and ViperGPT~\cite{viper} interleave reasoning with external tool execution, while recent methods including DeepEyesV2~\cite{hong2025deepeyesv2} and VisTA~\cite{huang2025visualtoolagent} employ reinforcement learning to optimize tool invocation strategies. However, these approaches target general-purpose reasoning and rely on visual feedback such as cropping or region enlargement. LocalAgent instead delegates suspicious regions to the LEV toolkit, comprising LAC and LGC, which perform invariant matching directly on raw visual features and return structured, interpretable evidence for closed-loop decision-making. A complementary auxiliary-capped RFT strategy further enables selective tool invocation without hand-crafted triggering rules.

\begin{figure}[t]
\centering
\includegraphics[width=1\linewidth]{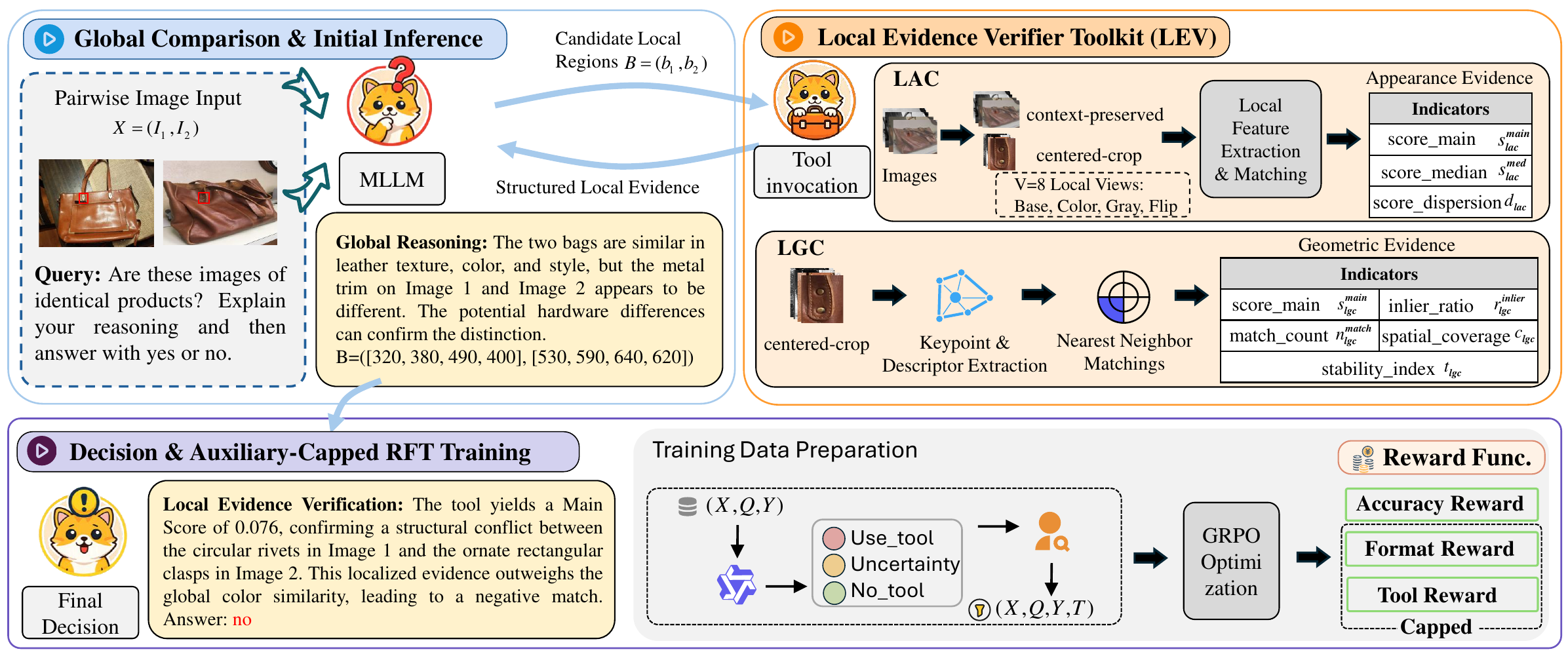}
\caption{
Overview of the LocalAgent framework. Given an image pair, the MLLM performs initial global semantic perception and either commits to a direct answer or, when the assessment is inconclusive, acts as a spatial controller to localize suspicious regions. These regions are conditionally delegated to the LEV toolkit, which returns structured scalar evidence that the MLLM integrates with its initial reasoning state to produce the final prediction. The strategic verification policy is optimized via auxiliary-capped RFT.
}
\label{fig:framework}
\end{figure}
\section{Methodology}
\label{sec:method}

\subsection{Overview}
\label{sec:overview}

LocalAgent decomposes instance-level verification into three roles (Fig.~\ref{fig:framework}). The MLLM acts as a spatial controller responsible for global semantic perception and the localization of suspicious regions. The Local Evidence Verifier (LEV) toolkit operates on vision-only features and returns structured scalar indicators. An RFT-trained policy governs whether verification is invoked at all. This division allocates each subtask to the component best suited to it.

\subsection{The MLLM as Spatial Controller}
Within LocalAgent, the MLLM is not bound to a fixed output role. Given an input image pair, the MLLM first performs global semantic perception to form an initial reasoning state. When this assessment is sufficient, the MLLM commits directly to a final answer, consistent with its strength on holistic understanding. When the global assessment is inconclusive, typically when two instances share strong semantic similarity yet may differ in localized details, the MLLM acts as a spatial controller and outputs a pair of region coordinates $B = (b_1, b_2)$, specifying where verification should be performed. In practice, the MLLM is prompted under a fixed instruction template to emit $B$ as normalized \texttt{[$x_1, y_1, x_2, y_2$]} coordinates within \texttt{<tool\_call>} tags.

This division aligns with two distinct visual abilities. Global semantic perception and region localization rely on semantic encoding, whereas deciding whether two localized regions correspond to the same physical instance requires preserving exactly the cues that semantic compression discards, and is therefore delegated to LEV (Sec.~\ref{sec:lev}). After LEV returns structured evidence, the MLLM integrates it with the initial reasoning state to produce the final prediction, closing the loop without re-feeding raw visual content.

\subsection{Local Evidence Verifier Toolkit}
\label{sec:lev}

The LEV toolkit supplies the discriminative evidence required when global semantic perception alone is insufficient. Matching is performed on features from encoders trained purely on visual signals, rather than on representations aligned with language. Self-supervised objectives at the patch level (DINO~\cite{dino}) and keypoint-descriptor learning (XFeat~\cite{potje2024xfeat}) preserve spatial and textural detail that can be attenuated by cross-modal alignment. LEV thus bypasses the semantic compression that motivates this work. We instantiate LEV with two sub-modules targeting complementary failure modes. DINO and XFeat are chosen as toolkit for their respective categories, representative. LAC is built upon DINO and performs patch-level appearance matching that remains stable under photometric variations where local structure is preserved despite changes in surface coloring. LGC is built upon XFeat and performs keypoint-based geometric matching that remains effective under viewpoint changes where appearance breaks but spatial correspondence is recoverable. The contribution of LEV lies not in the choice of backbones but in the unified scalar evidence interface that surfaces their outputs to the MLLM, enabling integration through evidential reasoning.

\subsubsection{Local Appearance Consistency (LAC)}

Given the localized regions $B$, LAC builds $V = 8$ view variants per region so that its estimate is less sensitive to background and photometric changes. For each region, it generates a context view that highlights the region within the full image and a crop view that isolates it from the background. Each view is kept in its original form and is also transformed by color jittering, grayscale conversion, and horizontal flipping, which gives four variants per view. For each view $v$, we extract patch-level DINO features and establish patch correspondences by cosine similarity, which yields the consistency ratio
\begin{equation}
s_v = \frac{|M^{(v)}|}{N^{(v)}},
\end{equation}
where $|M^{(v)}|$ denotes the number of stable patch-level matches and $N^{(v)}$ represents the total number of patches. To provide LocalAgent with interpretable feedback, LAC aggregates these scores into a structured evidence triplet $z_{\text{lac}} = (s_{\text{lac}}^{\text{main}}, s_{\text{lac}}^{\text{med}}, d_{\text{lac}})$. Specifically, the main appearance score
($s_{\text{lac}}^{\text{main}}$) is the mean of the top-$k$ scores, reflecting similarity under the most favorable views. The median appearance score ($s_{\text{lac}}^{\text{med}}$) captures the overall central tendency, and the view dispersion ($d_{\text{lac}}$) is the standard deviation across all views, reflecting the stability of the evidence. This compact representation allows the MLLM to perform evidential reasoning over explicit similarity indicators rather than high-dimensional raw features.

\subsubsection{Local Geometric Correspondence (LGC)}

The Local Geometric Correspondence (LGC) module characterizes spatial structural consistency between localized regions, providing robustness against drastic viewpoint and geometric deformations. Unlike LAC, the geometric estimation in LGC is less sensitive to color distribution. Therefore, the construction of view variants prioritizes perturbations in contrast and orientation by applying three fixed transformations, namely base, contrast equalization, and horizontal flip. 

For each view $v$, XFeat~\cite{potje2024xfeat} as a lightweight local feature extractor is utilized to identify keypoints and descriptors. A set of candidate matches $C^{(v)}$ is established based on descriptor similarity, which is subsequently filtered into an inlier set $I^{(v)} \subseteq C^{(v)}$ via robust geometric estimation. This process suppresses unstable correspondences induced by background clutter or repetitive textures. The single-view geometric score is defined as
\begin{equation}
g_v = \frac{|I^{(v)}|}{|C^{(v)}|},
\end{equation}
representing the proportion of matches supported by a consistent spatial relationship. To provide LocalAgent with comprehensive structural feedback, LGC aggregates these scores into a five-dimensional evidence vector $z_{\text{lgc}} = (s_{\text{lgc}}^{\text{main}}, r_{\text{lgc}}^{\text{inlier}}, n_{\text{lgc}}^{\text{match}}, c_{\text{lgc}}, t_{\text{lgc}})$. Specifically, the main geometric score ($s_{\text{lgc}}^{\text{main}}$) reflects the overall correspondence strength across all views, and the inlier ratio ($r_{\text{lgc}}^{\text{inlier}}$) measures matching precision. The effective match count ($n_{\text{lgc}}^{\text{match}}$) prevents reliance on sparse points. The spatial coverage ($c_{\text{lgc}}$) quantifies the distribution of matches to distinguish point-wise versus structural alignment, and the geometric stability ($t_{\text{lgc}}$) reflects evidence consistency under varying conditions.

\subsection{Learning When to Verify}
\label{sec:rft}

Many samples are resolvable through global semantic perception alone, whereas unconditional LEV invocation incurs unnecessary overhead. The decision of when to invoke verification thus becomes a learned behavior in itself, one that hand-crafted rules cannot easily capture across heterogeneous instance distributions. We formulate this as a strategic policy optimized via Group Relative Policy Optimization (GRPO)-based RFT~\cite{deepseekmath,deepseekr1}, with the reward signal dominated by final answer correctness and shaped by auxiliary signals on tool-use behavior and output format.

\paragraph{Weak behavioral priors.}

To stabilize early exploration and prevent data leakage, we construct weak behavioral priors on the training split using only its ground-truth labels. For each sample, the base model is evaluated under two zero-shot conditions, without tool access and with the actual LEV outputs supplied as reference feedback. A sample is labeled \texttt{No\_tool} when the model answers correctly without tools, and \texttt{Use\_tool} when it answers correctly only with tools. A held-out fraction of both categories is relabeled as \texttt{Uncertainty}, which breaks the deterministic \texttt{Use\_tool}/\texttt{No\_tool} boundary and serves as label-smoothing regularization for tool-use behavior. The labels shape only the auxiliary reward $R_{\text{tool}}$ and never enter $R_{\text{acc}}$, which remains supervised exclusively by the original ground-truth answers.

\paragraph{Auxiliary-capped reward.}
The reward combines an answer-correctness term $R_{\text{acc}}$, a tool-use term $R_{\text{tool}}$ aligned with the prior label $c \in \{\texttt{Use\_tool},\, \texttt{No\_tool},\, \texttt{Uncertainty}\}$, and a format term $R_{\text{fmt}}$ enforcing well-formed \texttt{<tool\_call>} and \texttt{<answer>} tags. $R_{\text{tool}}$ asymmetrically encourages verification on \texttt{Use\_tool} samples and restraint on \texttt{No\_tool} samples without dictating fixed rules. A valid tool call receives a positive reward on \texttt{Use\_tool} samples, a small penalty on \texttt{No\_tool} samples, and an intermediate positive reward on \texttt{Uncertainty} samples. Not invoking the verifier reverses the sign on the first two groups and is neutral on the third. Failed, repeated, or malformed calls receive the lowest tool reward regardless of the label. A naive linear combination of the three terms, however, would allow the policy to accumulate auxiliary rewards even when the final answer is wrong, yielding degenerate strategies that prioritize tool-use behavior over discriminative accuracy. The auxiliary contribution is therefore bounded whenever the prediction $\hat{y}$ is incorrect

\begin{equation}
r =
\begin{cases}
\alpha R_{\text{acc}} + \beta R_{\text{tool}} + \gamma R_{\text{fmt}}, & \hat{y} = y, \\
\alpha R_{\text{acc}} + \max\!\left(\min\!\left(\beta R_{\text{tool}} + \gamma R_{\text{fmt}},\, \delta\right),\, -\epsilon\right), & \hat{y} \neq y,
\end{cases}
\label{eq:reward}
\end{equation}
where $\delta$ and $\epsilon$ are small positive constants. The cap $\delta$ prevents the policy from accumulating positive auxiliary reward on wrong answers, while the floor $-\epsilon$ bounds the auxiliary penalty so that it does not outweigh the answer term. Within these bounds, negative auxiliary penalties still discourage degenerate tool-use behavior.

\begin{table*}[t]
\renewcommand{\arraystretch}{1.2}
\centering
\caption{
Main quantitative results (Accuracy, \%). P, M, and T denote Pair, Mul, and Total accuracy. The leading Mean, Pair, and Mul columns are the unweighted averages over the six data sources. $\dagger$ indicates closed-source models. Bold marks the best open-source result in each column.
}
\label{tab:main}
\resizebox{\textwidth}{!}{%
\setlength{\tabcolsep}{3.5pt}
\begin{tabular}{l ccc ccc ccc ccc ccc ccc ccc}
\toprule
\multirow{2}{*}{\textbf{Method}}
  & \multirow{2}{*}{\textbf{Mean}}
  & \multirow{2}{*}{\textbf{Pair}}
  & \multirow{2}{*}{\textbf{Mul}}
  & \multicolumn{3}{c}{\textbf{TWIN-Eval}}
  & \multicolumn{3}{c}{\textbf{ILIAS}}
  & \multicolumn{3}{c}{\textbf{Landmarks}}
  & \multicolumn{3}{c}{\textbf{MET}}
  & \multicolumn{3}{c}{\textbf{CUB}}
  & \multicolumn{3}{c}{\textbf{Inquire}} \\
\cmidrule(lr){5-7}\cmidrule(lr){8-10}\cmidrule(lr){11-13}
\cmidrule(lr){14-16}\cmidrule(lr){17-19}\cmidrule(lr){20-22}
  & & &
  & P & M & T & P & M & T & P & M & T
  & P & M & T & P & M & T & P & M & T \\
\midrule
\multicolumn{22}{l}{\textit{Closed-source Models}} \\
GPT-4o$^\dagger$
  & 86.1 & 91.1 & 81.1
  & 92.2 & 71.2 & 81.7
  & 95.0 & 95.5 & 95.3
  & 87.3 & 74.7 & 81.0
  & 86.3 & 72.9 & 79.6
  & 91.7 & 86.0 & 88.9
  & 94.3 & 86.3 & 90.3 \\
Gemini2.5 Flash$^\dagger$
  & 82.2 & 87.0 & 77.5
  & 89.1 & 68.1 & 78.6
  & 92.2 & 90.5 & 91.4
  & 85.4 & 77.0 & 81.2
  & 82.6 & 71.9 & 77.3
  & 84.1 & 77.9 & 81.0
  & 88.3 & 79.4 & 83.9 \\
\midrule
\multicolumn{22}{l}{\textit{Open-source Baselines}} \\
PerceptionLM-3B
  & 37.8 & 50.0 & 25.5
  & 46.9 & 25.2 & 36.1
  & 49.6 & 25.0 & 37.3
  & 51.5 & 25.7 & 38.6
  & 50.4 & 26.1 & 38.3
  & 49.6 & 24.9 & 37.3
  & 52.1 & 26.1 & 39.1 \\
Moondream2-2B
  & 45.9 & 61.6 & 30.2
  & 64.5 & 30.0 & 47.3
  & 58.7 & 30.3 & 44.5
  & 59.8 & 25.2 & 42.5
  & 84.3 & 27.0 & 55.7
  & 50.6 & 43.3 & 47.0
  & 51.9 & 25.1 & 38.5 \\
Gemma3-4B
  & 68.3 & 86.1 & 50.4
  & 68.1 & 39.5 & 53.8
  & 88.4 & 46.5 & 67.5
  & \textbf{91.2} & 55.2 & 73.2
  & \textbf{87.9} & 57.5 & 72.7
  & 91.2 & 55.2 & 73.2
  & 90.0 & 48.7 & 69.4 \\
SmolVLM2-2.2B
  & 46.9 & 64.9 & 28.8
  & 54.7 & 25.1 & 39.9
  & 52.2 & 29.7 & 41.0
  & 54.1 & 29.6 & 41.9
  & 73.9 & 27.9 & 50.9
  & 81.5 & 35.0 & 58.3
  & 73.0 & 25.7 & 49.4 \\
Qwen3-VL-4B
  & 73.3 & 76.8 & 69.7
  & \textbf{89.1} & \textbf{76.7} & \textbf{82.9}
  & \textbf{89.2} & \textbf{79.3} & \textbf{84.3}
  & 50.9 & 60.1 & 55.5
  & 54.0 & 50.1 & 52.1
  & 89.6 & 77.5 & 83.6
  & 87.9 & 74.6 & 81.3 \\
\midrule
\multicolumn{22}{l}{\textit{RFT Baselines}} \\
CoT (Qwen3-VL-4B)
  & 75.8 & 83.4 & 68.1
  & 86.6 & 66.4 & 76.5
  & 84.4 & 75.8 & 80.1
  & 75.7 & 52.9 & 64.3
  & 76.9 & 58.9 & 67.9
  & 84.4 & 79.7 & 82.1
  & 92.3 & 75.0 & 83.7 \\
TWIN
  & 67.0 & 86.7 & 47.2
  & 88.9 & 45.7 & 67.3
  & 79.9 & 43.7 & 61.8
  & 80.8 & 34.9 & 57.9
  & 85.4 & 46.5 & 66.0
  & 92.5 & 57.7 & 75.1
  & 92.7 & 54.6 & 73.7 \\
\midrule
\multicolumn{22}{l}{\textit{Ours}} \\
\rowcolor{oursrow}
\textbf{LocalAgent}
  & \textbf{80.7} & \textbf{87.6} & \textbf{73.9}
  & 80.5 & 68.5 & 74.5
  & 82.7 & 76.4 & 79.6
  & 87.3 & \textbf{65.0} & \textbf{76.2}
  & 86.7 & \textbf{71.9} & \textbf{79.3}
  & \textbf{93.2} & \textbf{82.5} & \textbf{87.9}
  & \textbf{94.9} & \textbf{78.8} & \textbf{86.9} \\
\bottomrule
\end{tabular}%
}
\end{table*}

\section{Experiments}
\label{sec:experiments}

\subsection{Experimental Setup}

\paragraph{Train Data and Protocols.}
We conduct experiments following the evaluation sets and protocols established by the TWIN benchmark~\cite{twin}. As a large-scale VQA dataset designed for fine-grained visual perception, TWIN includes positive, negative, and hard negative pairs. TWIN is constructed from product images sourced from Amazon Reviews, with visually similar hard negatives selected and manually screened during dataset construction. For the RFT stage, we select training samples with an emphasis on hard negative pairs with very high visual similarity and on positive pairs with large changes in viewpoint, illumination, or background.

\paragraph{Benchmarks and Evaluation Metrics.} The evaluation suite spans six distinct data sources to ensure robust generalization, namely TWIN-Eval for in-distribution testing, ILIAS \cite{ilias} for retail products, Landmarks \cite{landmarks} for geographical sites, MET \cite{met} for artwork, CUB \cite{cub} for birds, and Inquire \cite{inquire} for flora and fauna. In total, 12k queries across these fine-grained domains are evaluated. We utilize two task formats for assessment. The pairwise task requires a binary judgment on whether two images correspond to the same instance. The multiple-choice task presents a query image alongside three candidate images and asks the model to determine how many candidates depict the same physical instance as the query. Performance is quantified using mean accuracy alongside specific metrics for each task format across the six sources.

\paragraph{General Capability Benchmarks.} To ensure that the RFT process does not compromise general multimodal and linguistic capabilities, evaluations are performed on two sets of general benchmarks. Vision-language benchmarks include MME-RealWorld-Lite~\cite{mmereallworld}, AI2D~\cite{ai2d}, SEED~\cite{seed}, ChartQA~\cite{chartqa}, TextVQA~\cite{textvqa}, MMMU~\cite{mmmu}, and BLINK~\cite{blink}. Text-only benchmarks include MMLU~\cite{mmlu}, HellaSwag~\cite{hellaswag}, and GSM8K~\cite{gsm8k}.

\paragraph{Baselines.} The evaluation includes a broad spectrum of comparison methods. We select closed-source models including GPT-4o~\cite{gpt4o} and Gemini2.5 Flash~\cite{gemini} as references for frontier performance. Open-source baselines comprise models of similar parameter scales such as Qwen3-VL-4B~\cite{qwen_vl_rft}, PerceptionLM-3B~\cite{perceptionlm}, Moondream2-2B~\cite{moondream2}, Gemma3-4B~\cite{gemma3}, and SmolVLM2-2.2B~\cite{smolvlm2}. We also compare against the TWIN model~\cite{twin} and a CoT-RFT baseline, denoted CoT (Qwen3-VL-4B) in the tables, which is trained on the same dataset as LocalAgent using RFT with accuracy and format rewards and without any external tools.

\subsection{Main Results}

Table~\ref{tab:main} presents the comparative results across the six data sources. LocalAgent improves mean accuracy by 4.9\% over the strongest RFT baseline CoT (Qwen3-VL-4B) while invoking a verifier on only 12.13\% of samples. Paired-bootstrap confidence intervals and McNemar tests confirm that this improvement is statistically significant (Appendix~\ref{appendix:statistical}). These results further narrow the gap with closed-source frontiers like Gemini2.5 Flash. LocalAgent improves both the mean pairwise and the mean multiple-choice accuracy. On TWIN-Eval, TWIN reaches 88.9\% accuracy on pairwise queries but 45.7\% on multiple-choice queries. In contrast, LocalAgent maintains balanced performance across the two formats. On TWIN-Eval, however, LocalAgent remains below the untrained base model, indicating that the current RFT policy and selective verification strategy do not fully match this benchmark's evaluation distribution. We therefore treat this benchmark-specific regression as a limitation of the current training setup rather than attributing it to a single mechanism. LocalAgent improves over CoT-RFT on four of the five out-of-distribution sources, while its mixed performance on ILIAS and TWIN-Eval indicates that the current training and selective-verification setup does not transfer uniformly across evaluation distributions.

\begin{table*}[t]
\centering

\begin{minipage}[t]{0.49\textwidth}
\centering
\caption{
    Ablation study on the LEV toolkit. Pair/Mul/All Call report the tool-invocation rates for each variant.
}
\label{tab:tool_ablation}
\small
\setlength{\tabcolsep}{4pt}
\resizebox{\textwidth}{!}{%
\begin{tabular}{l ccc ccc}
\toprule
\multirow{2}{*}{\textbf{Method}}
  & \multicolumn{3}{c}{\textbf{Performance (\%)}}
  & \multicolumn{3}{c}{\textbf{Invocation Rate (\%)}} \\
\cmidrule(lr){2-4}\cmidrule(lr){5-7}
  & \textbf{Mean} & \textbf{Pair} & \textbf{Mul}
  & \textbf{Pair} & \textbf{Mul} & \textbf{All} \\
\midrule
Only LAC   & 78.82 & 85.45 & 72.18 & 31.80 & 10.77 & 21.29 \\
Only LGC   & 79.94 & \textbf{88.57} & 71.32 & 30.47 &  8.53 & 19.50 \\
\rowcolor{oursrow}
\textbf{LAC+LGC (Ours)}
          & \textbf{80.70} & 87.55 & \textbf{73.85}
          & 7.85 & 16.40 & 12.13 \\
\bottomrule
\end{tabular}%
}
\end{minipage}
\hfill
\begin{minipage}[t]{0.49\textwidth}
\centering
\caption{Ablation results of the reward function components and gating mechanisms on instance level verification tasks.}
\label{tab:reward_ablation}
\small
\setlength{\tabcolsep}{4pt}
\resizebox{\textwidth}{!}{%
\begin{tabular}{l ccc ccc}
\toprule
\multirow{2}{*}{\textbf{Method}}
  & \multicolumn{3}{c}{\textbf{Performance (\%)}}
  & \multicolumn{3}{c}{\textbf{Invocation Rate (\%)}} \\
\cmidrule(lr){2-4}\cmidrule(lr){5-7}
  & \textbf{Mean} & \textbf{Pair} & \textbf{Mul}
  & \textbf{Pair} & \textbf{Mul} & \textbf{All} \\
\midrule
w/o weak prior & 79.39 & 86.55 & 72.23 & 38.45 & 22.90 & 30.68 \\
Hard gating    & 80.23 & \textbf{87.72} & 72.73 & 10.58 & 17.83 & 14.21 \\
\rowcolor{oursrow}
\textbf{LocalAgent (Ours)}
               & \textbf{80.70} & 87.55 & \textbf{73.85}
               & 7.85 & 16.40 & 12.13 \\
\bottomrule
\end{tabular}%
}
\end{minipage}

\end{table*}

\paragraph{Comparison with crop-based verification.}
To test whether re-feeding localized crops is sufficient, we compare LocalAgent with Qwen3-VL-4B equipped with a crop tool that returns the image regions inside the predicted boxes, trained and evaluated under the same protocol. As shown in Table~\ref{tab:crop_baseline}, LocalAgent improves pairwise accuracy from 76.81\% to 87.55\% and mean accuracy from 75.64\% to 80.70\%, while its multiple-choice accuracy is slightly lower. Pairwise verification maps localized evidence to one binary decision. Multiple-choice verification also requires comparing evidence across three candidates, which the current policy does not jointly aggregate.

\subsection{Ablation Study}
\paragraph{Tool Ablation.}

To evaluate the distinct contributions of individual verification modules, we analyze the performance and invocation patterns of the LAC and LGC components in Table~\ref{tab:tool_ablation}. It shows that LAC and LGC each surpass the CoT-RFT baseline individually, while their integration reaches 80.70\% mean accuracy, suggesting the complementarity of appearance and geometric evidence. The coordinated policy also invokes LEV less often overall, on 12.13\% of samples compared with 21.29\% and 19.50\% for the single-tool variants. This decrease comes from the pairwise task, whereas the multiple-choice call rate rises from 10.77\% and 8.53\% to 16.40\%.

\paragraph{Reward Ablation.}
Table~\ref{tab:reward_ablation} examines two design choices. Without the weak prior, the pairwise invocation rate rises to 38.45\% and mean accuracy drops by 1.31\%, which suggests that the prior helps the policy identify samples that benefit from verification. Replacing the auxiliary cap with hard gating removes all positive auxiliary reward on incorrect predictions, leading to a 0.47\% accuracy drop. Both components therefore contribute to the final performance. A sensitivity analysis of the auxiliary bounds is given in Appendix~\ref{appendix:reward_sensitivity}.

\begin{table*}[t]
\centering
\begin{minipage}[t]{0.49\textwidth}
\centering
\caption{Accuracy (\%) of crop-based tool augmentation and LocalAgent on the pairwise, multiple-choice, and mean settings.\strut}
\label{tab:crop_baseline}
\footnotesize
\setlength{\tabcolsep}{4pt}
\begin{tabular}{l ccc}
\toprule
\textbf{Method} & \textbf{Pair} & \textbf{Mul} & \textbf{Mean} \\
\midrule
Qwen3-VL-4B + crop & 76.81 & \textbf{74.47} & 75.64 \\
\rowcolor{oursrow}
\textbf{LocalAgent (Ours)} & \textbf{87.55} & 73.85 & \textbf{80.70} \\
\bottomrule
\end{tabular}
\end{minipage}
\hfill
\begin{minipage}[t]{0.49\textwidth}
\centering
\caption{Tool invocation behavior of LocalAgent. Easy and Hard denote invocation rates (\%) on baseline-correct and baseline-incorrect samples.\strut}
\label{tab:diagnostic}
\footnotesize
\setlength{\tabcolsep}{4pt}
\begin{tabular}{l ccc}
\toprule
\textbf{Setting} & \textbf{Easy} & \textbf{Hard} & \textbf{Call Rate} \\
\midrule
Pair & 6.81 & 11.27 & 7.85 \\
Mul & 13.71 & 22.55 & 16.40 \\
\bottomrule
\end{tabular}
\end{minipage}
\end{table*}

\subsection{Diagnostic Analysis}
\label{sec:analysis}

\paragraph{Tool Invocation Behavior.}
To examine whether LocalAgent learns a genuinely selective verification policy rather than invoking tools indiscriminately, we analyze invocation behavior across task formats and sample difficulty. As shown in Table~\ref{tab:diagnostic}, the overall call rates remain low at 7.85\% and 16.40\% for pairwise and multiple-choice respectively, confirming that accuracy gains are not simply a product of increased tool usage. Across both settings, the call rate (hard) on baseline-incorrect samples is consistently higher than the call rate (easy) on baseline-correct. This difficulty-sensitive pattern indicates that the learned policy concentrates LEV invocations on harder cases across both task formats.

\begin{figure}[t]
\centering
\includegraphics[width=0.9\linewidth]{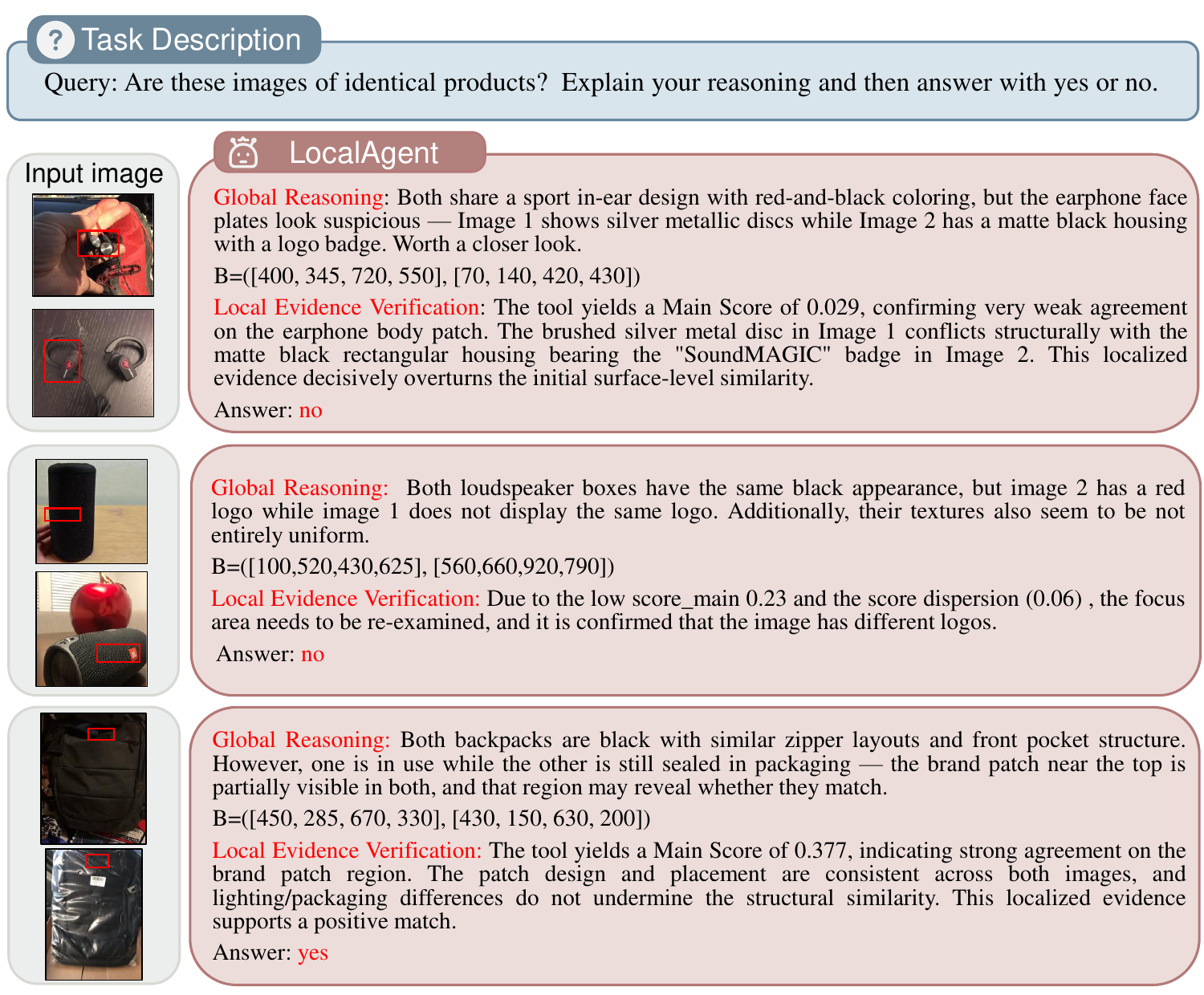}
\caption{
    Qualitative results of LocalAgent on pairwise verification cases.
}
\label{fig:qualitative}
\end{figure}

\paragraph{Inference Latency.}
Table~\ref{tab:latency_policy} reports policy-level end-to-end latency. The selective LAC+LGC policy invokes tools on 7.85\% of pairwise samples, 16.40\% of multiple-choice samples, and 12.13\% overall, keeping mean latency at 336.86 ms compared with 258.13 ms for direct inference. Selective invocation reduces mean latency from 610.41 ms/sample for the always-call policy to 336.86 ms/sample by avoiding unnecessary verifier executions, while the table makes the remaining per-sample overhead explicit.

\paragraph{Comparison with pure visual matching.}
To examine whether instance-level consistency verification can be solved by visual matching alone, we evaluate two non-MLLM baselines built on the LEV backbones, DINO and XFeat. Each baseline outputs a similarity score per sample, which is converted into a binary prediction with a threshold $t$. To avoid disadvantaging the baselines through an arbitrary choice of $t$, we sweep $t$ over a dense grid on each evaluation subset and report the accuracy under the optimal threshold, which gives an upper-bound estimate for a single-threshold rule. In full-image matching, the whole images are matched without localization, and for the multiple-choice task the query is compared with each of the three candidates and the highest-scoring one is selected. In MLLM-localized matching, the images are first cropped to the suspicious regions predicted by Qwen3-VL-4B, following the procedure in Appendix~\ref{appendix:localization}. As shown in Table~\ref{tab:matching_baselines}, MLLM-localized matching improves the mean accuracy of DINO from 74.23\% to 75.06\% and that of XFeat from 69.23\% to 71.11\%, indicating that the localized regions contain discriminative cues. Both baselines nevertheless remain well below LocalAgent, whose mean accuracy is 80.70\%. DINO with MLLM-localized matching slightly exceeds LocalAgent on the multiple-choice task, 75.36\% versus 73.85\%, but only with a threshold selected post hoc on the evaluation set, which is unavailable at deployment.

\begin{table*}[t]
\centering
\begin{minipage}[t]{0.49\textwidth}
\centering
\caption{Policy-level end-to-end inference latency under different invocation settings, reported in ms per sample on the pairwise and multiple-choice tasks.\strut}
\label{tab:latency_policy}
\footnotesize
\setlength{\tabcolsep}{4pt}
\begin{tabular}{l ccc}
\toprule
\textbf{Setting} & \textbf{Pair} & \textbf{Mul} & \textbf{Mean} \\
\midrule
Direct inference & \textbf{184.72} & \textbf{331.54} & \textbf{258.13} \\
Only LAC & 290.50 & 350.48 & 320.49 \\
Only LGC & 294.87 & 379.18 & 337.03 \\
\rowcolor{oursrow}
\textbf{LAC+LGC (Ours)} & 220.54 & 453.17 & 336.86 \\
Always-call policy & 400.53 & 820.29 & 610.41 \\
\bottomrule
\end{tabular}
\end{minipage}
\hfill
\begin{minipage}[t]{0.49\textwidth}
\centering
\caption{Accuracy (\%) of pure visual matching baselines under the optimal threshold, computed on the full images and on the regions localized by Qwen3-VL-4B.\strut}
\label{tab:matching_baselines}
\footnotesize
\setlength{\tabcolsep}{4pt}
\begin{tabular}{l ccc}
\toprule
\textbf{Method} & \textbf{Pair} & \textbf{Mul} & \textbf{Mean} \\
\midrule
DINO (full) & 73.18 & 75.28 & 74.23 \\
XFeat (full) & 68.96 & 69.50 & 69.23 \\
DINO (localized) & 74.75 & \textbf{75.36} & 75.06 \\
XFeat (localized) & 70.88 & 71.33 & 71.11 \\
\rowcolor{oursrow}
\textbf{LocalAgent (Ours)} & \textbf{87.55} & 73.85 & \textbf{80.70} \\
\bottomrule
\end{tabular}
\end{minipage}
\end{table*}

\begin{table}[!t]
\centering
\caption{General capability evaluation (\%). LocalAgent matches Qwen3-VL-4B across vision-language and text-only benchmarks.}
\label{tab:generalization}
\small
\resizebox{\textwidth}{!}{%
\begin{tabular}{l ccccccc ccc}
\toprule
& \multicolumn{7}{c}{\textbf{Vision-Language}} & \multicolumn{3}{c}{\textbf{Text-only}} \\
\cmidrule(lr){2-8} \cmidrule(lr){9-11}
\textbf{Method} & \textbf{MME-RW-L} & \textbf{AI2D} & \textbf{SEED} & \textbf{ChartQA} & \textbf{TextVQA} & \textbf{MMMU} & \textbf{BLINK} & \textbf{MMLU} & \textbf{HellaSwag} & \textbf{GSM8K} \\
\midrule
Qwen3-VL-4B          & 29.03 & 84.20 & 73.05 & 75.00 & 70.35 & 77.20 & 65.8 & 68.80 & 86.90 & 89.54 \\
\rowcolor{oursrow}
\textbf{LocalAgent}  & 29.39 & 84.44 & 73.28 & 75.08 & 70.39 & 77.22 & 65.9 & 68.84 & 86.96 & 89.54 \\
\bottomrule
\end{tabular}}
\end{table}

\subsection{General Capability Evaluation}

A primary concern in RFT for domain-specific tasks is the potential for catastrophic forgetting of the model's general capabilities~\cite{wu2025f,yan2025task}. We evaluate LocalAgent on ten general-purpose benchmarks spanning vision-language and text-only tasks. As shown in Table~\ref{tab:generalization}, LocalAgent matches or slightly exceeds the Qwen3-VL-4B baseline on every benchmark, indicating that our RFT improves fine-grained verification without degrading general capabilities.

\subsection{Qualitative Analysis}

Fig.~\ref{fig:qualitative} illustrates two complementary verification patterns. Low LAC main scores overturn deceptive global similarity, while a high main score confirms identity despite packaging-induced appearance shifts. These cases demonstrate that structured scalar evidence enables LocalAgent to resolve ambiguous cases beyond the reach of global semantic perception.

\section{Conclusion}
We present LocalAgent, a collaborative agentic framework for fine-grained instance-level consistency verification. The MLLM contributes global semantic perception and suspicious-region localization, while the LEV toolkit supplies appearance and geometric evidence from vision-only encoders, returned as structured scalar indicators that are integrated through textual reasoning rather than re-fed by the vision encoder. An auxiliary-capped RFT strategy further enables the policy to learn when verification is necessary. Experiments across six diverse data sources show that LocalAgent improves mean accuracy while preserving general multimodal capabilities, narrowing the gap to closed-source counterparts. Beyond instance-level verification, this paradigm of coupling MLLM-level semantic control with specialized non-semantic verifiers suggests a general recipe for tasks where vision-language alignment alone can attenuate the fine-grained signals required by the task.

\paragraph{Limitations.}
LEV examines only the regions proposed by the MLLM, so the quality of its evidence depends on the localization step. The current policy does not aggregate evidence jointly across the candidates of a multiple-choice query. In addition, LocalAgent remains below the base model on TWIN-Eval (Appendix~\ref{appendix:twin_regression}) and is 30.5\% slower than direct inference (Table~\ref{tab:latency_policy}).

\begin{ack}
This work was supported by the General Program of the National Natural Science Foundation of China (No. 62676278), the New Generation Artificial Intelligence---National Science and Technology Major Project (No. 2025ZD0123501), the National Natural Science Foundation of China (No. 62406320), the State Key Laboratory of Multimodal Artificial Intelligence Systems (No. MAIS20260206), the Open Research Fund of the State Key Laboratory of Blockchain and Data Security, Zhejiang University (No. A2542), and the AI Super Computing Platform of Macau University of Science and Technology.

The authors declare no competing interests.
\end{ack}

\bibliography{reference}

@inproceedings{hallusionbench,
  title={{HallusionBench}: an advanced diagnostic suite for entangled language hallucination and visual illusion in large vision-language models},
  author={Guan, Tianrui and Liu, Fuxiao and Wu, Xiyang and Xian, Ruiqi and Li, Zongxia and Liu, Xiaoyu and Wang, Xijun and Chen, Lichang and Huang, Furong and Yacoob, Yaser and others},
  booktitle={Proceedings of the IEEE/CVF conference on computer vision and pattern recognition},
  pages={14375--14385},
  year={2024}
}

@inproceedings{blink,
  title={{BLINK}: Multimodal large language models can see but not perceive},
  author={Fu, Xingyu and Hu, Yushi and Li, Bangzheng and Feng, Yu and Wang, Haoyu and Lin, Xudong and Roth, Dan and Smith, Noah A and Ma, Wei-Chiu and Krishna, Ranjay},
  booktitle={European Conference on Computer Vision},
  pages={148--166},
  year={2024}
}

@inproceedings{mmvp,
  title={Eyes wide shut? {E}xploring the visual shortcomings of multimodal {LLMs}},
  author={Tong, Shengbang and Liu, Zhuang and Zhai, Yuexiang and Ma, Yi and LeCun, Yann and Xie, Saining},
  booktitle={Proceedings of the IEEE/CVF conference on computer vision and pattern recognition},
  pages={9568--9578},
  year={2024}
}

@inproceedings{pope,
  title={Evaluating object hallucination in large vision-language models},
  author={Li, Yifan and Du, Yifan and Zhou, Kun and Wang, Jinpeng and Zhao, Xin and Wen, Ji-Rong},
  booktitle={Proceedings of the 2023 conference on empirical methods in natural language processing},
  pages={292--305},
  year={2023}
}

@inproceedings{mme,
  title={{MME}: A Comprehensive Evaluation Benchmark for Multimodal Large Language Models},
  author={Fu, Chaoyou and Chen, Peixian and Shen, Yunhang and Qin, Yulei and Zhang, Mengdan and Lin, Xu and Yang, Jinrui and Zheng, Xiawu and Li, Ke and Sun, Xing and others},
  booktitle={The Thirty-ninth Annual Conference on Neural Information Processing Systems Datasets and Benchmarks Track}
}

@article{mmreact,
  title={{MM-REACT}: Prompting {ChatGPT} for multimodal reasoning and action},
  author={Yang, Zhengyuan and Li, Linjie and Wang, Jianfeng and Lin, Kevin and Azarnasab, Ehsan and Ahmed, Faisal and Liu, Zicheng and Liu, Ce and Zeng, Michael and Wang, Lijuan},
  journal={arXiv preprint arXiv:2303.11381},
  year={2023}
}

@inproceedings{viper,
  title={{ViperGPT}: Visual inference via {Python} execution for reasoning},
  author={Sur{\'\i}s, D{\'\i}dac and Menon, Sachit and Vondrick, Carl},
  booktitle={Proceedings of the IEEE/CVF international conference on computer vision},
  pages={11888--11898},
  year={2023}
}

@inproceedings{visuallookback,
  title={Look-back: Implicit visual re-focusing in {MLLM} reasoning},
  author={Yang, Shuo and Niu, Yuwei and Liu, Yuyang and Ye, Yang and Lin, Bin and Yuan, Li},
  booktitle={Proceedings of the AAAI Conference on Artificial Intelligence},
  pages={11694--11702},
  year={2026}
}

@article{deepseekmath,
  title={{DeepSeekMath}: Pushing the limits of mathematical reasoning in open language models},
  author={Shao, Zhihong and Wang, Peiyi and Zhu, Qihao and Xu, Runxin and Song, Junxiao and Bi, Xiao and Zhang, Haowei and Zhang, Mingchuan and Li, YK and Wu, Yang and others},
  journal={arXiv preprint arXiv:2402.03300},
  year={2024}
}

@article{qwen_vl_rft,
  title={{Qwen3-VL} technical report},
  author={Bai, Shuai and Cai, Yuxuan and Chen, Ruizhe and Chen, Keqin and Chen, Xionghui and Cheng, Zesen and Deng, Lianghao and Ding, Wei and Gao, Chang and Ge, Chunjiang and others},
  journal={arXiv preprint arXiv:2511.21631},
  year={2025}
}

@article{huang2025visualtoolagent,
  title={{VisualToolAgent} ({VisTA}): A reinforcement learning framework for visual tool selection},
  author={Huang, Zeyi and Ji, Yuyang and Rajan, Anirudh Sundara and Cai, Zefan and Xiao, Wen and Wang, Haohan and Hu, Junjie and Lee, Yong Jae},
  journal={arXiv preprint arXiv:2505.20289},
  year={2025}
}

@article{hong2025deepeyesv2,
  title={{DeepEyesV2}: Toward agentic multimodal model},
  author={Hong, Jack and Zhao, Chenxiao and Zhu, ChengLin and Lu, Weiheng and Xu, Guohai and Yu, Xing},
  journal={arXiv preprint arXiv:2511.05271},
  year={2025}
}

@inproceedings{twin,
  title={Same or Not? {E}nhancing Visual Perception in Vision-Language Models},
  author={Marsili, Damiano and Mehta, Aditya and Lin, Ryan Y and Gkioxari, Georgia},
  booktitle={Proceedings of the Computer Vision and Pattern Recognition Conference},
  year={2026}
}

@article{deepseekr1,
   title={{DeepSeek-R1} incentivizes reasoning in {LLMs} through reinforcement learning},
   volume={645},
   number={8081},
   journal={Nature},
    author={Guo, Daya and Yang, Dejian and Zhang, Haowei and Song, Junxiao and Wang, Peiyi and Zhu, Qihao and Xu, Runxin and Zhang, Ruoyu and Ma, Shirong and Bi, Xiao and others},
   year={2025},
   pages={633–638} }

@inproceedings{mmereallworld,
  title={{MME-RealWorld}: Could your multimodal {LLM} challenge high-resolution real-world scenarios that are difficult for humans?},
  author={Zhang, YiFan and Zhang, Huanyu and Tian, Haochen and Fu, Chaoyou and Zhang, Shuangqing and Wu, Junfei and Li, Feng and Wang, Kun and Wen, Qingsong and Zhang, Zhang and others},
  booktitle={International Conference on Learning Representations},
  volume={2025},
  pages={89655--89701},
  year={2025}
}

@inproceedings{ai2d,
  title={A diagram is worth a dozen images},
  author={Kembhavi, Aniruddha and Salvato, Mike and Kolve, Eric and Seo, Minjoon and Hajishirzi, Hannaneh and Farhadi, Ali},
  booktitle={European conference on computer vision},
  pages={235--251},
  year={2016},
}

@article{seed,
  title={{SEED-Bench}: Benchmarking multimodal {LLMs} with generative comprehension},
  author={Li, Bohao and Wang, Rui and Wang, Guangzhi and Ge, Yuying and Ge, Yixiao and Shan, Ying},
  journal={arXiv preprint arXiv:2307.16125},
  year={2023}
}

@inproceedings{mmmu,
  title={{MMMU}: A massive multi-discipline multimodal understanding and reasoning benchmark for expert {AGI}},
  author={Yue, Xiang and Ni, Yuansheng and Zhang, Kai and Zheng, Tianyu and Liu, Ruoqi and Zhang, Ge and Stevens, Samuel and Jiang, Dongfu and Ren, Weiming and Sun, Yuxuan and others},
  booktitle={Proceedings of the IEEE/CVF conference on computer vision and pattern recognition},
  pages={9556--9567},
  year={2024}
}

@inproceedings{chartqa,
  title={{ChartQA}: A benchmark for question answering about charts with visual and logical reasoning},
  author={Masry, Ahmed and Do, Xuan Long and Tan, Jia Qing and Joty, Shafiq and Hoque, Enamul},
  booktitle={Findings of the association for computational linguistics: ACL 2022},
  pages={2263--2279},
  year={2022}
}

@inproceedings{textvqa,
  title={Towards {VQA} models that can read},
  author={Singh, Amanpreet and Natarajan, Vivek and Shah, Meet and Jiang, Yu and Chen, Xinlei and Batra, Dhruv and Parikh, Devi and Rohrbach, Marcus},
  booktitle={Proceedings of the IEEE/CVF conference on computer vision and pattern recognition},
  pages={8317--8326},
  year={2019}
}

@inproceedings{mmlu,
  title={Measuring Massive Multitask Language Understanding},
  author={Hendrycks, Dan and Burns, Collin and Basart, Steven and Zou, Andy and Mazeika, Mantas and Song, Dawn and Steinhardt, Jacob},
  booktitle={International Conference on Learning Representations}
}

@inproceedings{hellaswag,
  title={{HellaSwag}: Can a machine really finish your sentence?},
  author={Zellers, Rowan and Holtzman, Ari and Bisk, Yonatan and Farhadi, Ali and Choi, Yejin},
  booktitle={Proceedings of the 57th annual meeting of the association for computational linguistics},
  pages={4791--4800},
  year={2019}
}

@article{gsm8k,
  title={Training verifiers to solve math word problems},
  author={Cobbe, Karl and Kosaraju, Vineet and Bavarian, Mohammad and Chen, Mark and Jun, Heewoo and Kaiser, Lukasz and Plappert, Matthias and Tworek, Jerry and Hilton, Jacob and Nakano, Reiichiro and others},
  journal={arXiv preprint arXiv:2110.14168},
  year={2021}
}

@inproceedings{met,
  title={The {MET} dataset: Instance-level recognition for artworks},
  author={Ypsilantis, Nikolaos-Antonios and Garcia, Noa and Han, Guangxing and Ibrahimi, Sarah and Van Noord, Nanne and Tolias, Giorgos},
  booktitle={Thirty-fifth conference on neural information processing systems datasets and benchmarks track (Round 2)},
  year={2021}
}

@inproceedings{ilias,
  title={{ILIAS}: Instance-level image retrieval at scale},
  author={Kordopatis-Zilos, Giorgos and Stojni{\'c}, Vladan and Manko, Anna and Suma, Pavel and Ypsilantis, Nikolaos-Antonios and Efthymiadis, Nikos and Laskar, Zakaria and Matas, Jiri and Chum, Ondrej and Tolias, Giorgos},
  booktitle={Proceedings of the Computer Vision and Pattern Recognition Conference},
  pages={14777--14787},
  year={2025}
}

@inproceedings{landmarks,
  title={Google landmarks dataset v2-a large-scale benchmark for instance-level recognition and retrieval},
  author={Weyand, Tobias and Araujo, Andre and Cao, Bingyi and Sim, Jack},
  booktitle={Proceedings of the IEEE/CVF conference on computer vision and pattern recognition},
  pages={2575--2584},
  year={2020}
}

@article{cub,
  title={The {Caltech-UCSD} birds-200-2011 dataset},
  author={Wah, Catherine and Branson, Steve and Welinder, Peter and Perona, Pietro and Belongie, Serge},
  year={2011}
}

@article{inquire,
  title={{INQUIRE}: A natural world text-to-image retrieval benchmark},
  author={Vendrow, Edward and Pantazis, Omiros and Shepard, Alexander and Brostow, Gabriel and Jones, Kate E and Mac Aodha, Oisin and Beery, Sara and Van Horn, Grant},
  journal={Advances in Neural Information Processing Systems},
  volume={37},
  pages={126500--126514},
  year={2024}
}

@article{gpt4o,
  title={{GPT-4o} system card},
  author={Hurst, Aaron and Lerer, Adam and Goucher, Adam P and Perelman, Adam and Ramesh, Aditya and Clark, Aidan and Ostrow, AJ and Welihinda, Akila and Hayes, Alan and Radford, Alec and others},
  journal={arXiv preprint arXiv:2410.21276},
  year={2024}
}

@article{gemini,
  title={Gemini 2.5: Pushing the frontier with advanced reasoning, multimodality, long context, and next generation agentic capabilities},
  author={Comanici, Gheorghe and Bieber, Eric and Schaekermann, Mike and Pasupat, Ice and Sachdeva, Noveen and Dhillon, Inderjit and Blistein, Marcel and Ram, Ori and Zhang, Dan and Rosen, Evan and others},
  journal={arXiv preprint arXiv:2507.06261},
  year={2025}
}

@article{perceptionlm,
  title={{PerceptionLM}: Open-access data and models for detailed visual understanding},
  author={Cho, Jang Hyun and Madotto, Andrea and Mavroudi, Effrosyni and Afouras, Triantafyllos and Nagarajan, Tushar and Maaz, Muhammad and Song, Yale and Ma, Tengyu and Hu, Shuming and Jain, Suyog and others},
  journal={arXiv preprint arXiv:2504.13180},
  year={2025}
}

@article{gemma3,
  title={Gemma 3 technical report},
  author={Team, Gemma and Kamath, Aishwarya and Ferret, Johan and Pathak, Shreya and Vieillard, Nino and Merhej, Ramona and Perrin, Sarah and Matejovicova, Tatiana and Ram{\'e}, Alexandre and Rivi{\`e}re, Morgane and others},
  journal={arXiv preprint arXiv:2503.19786},
  year={2025}
}

@article{smolvlm2,
  title={{SmolVLM}: Redefining small and efficient multimodal models},
  author={Marafioti, Andr{\'e}s and Zohar, Orr and Farr{\'e}, Miquel and Noyan, Merve and Bakouch, Elie and Cuenca, Pedro and Zakka, Cyril and Allal, Loubna Ben and Lozhkov, Anton and Tazi, Nouamane and others},
  journal={arXiv preprint arXiv:2504.05299},
  year={2025}
}

@software{moondream2,
  author  = {{Moondream}},
  title   = {moondream2},
  year    = {2025},
  url     = {https://github.com/m87-labs/moondream}
}

@inproceedings{dino,
  title={Emerging properties in self-supervised vision transformers},
  author={Caron, Mathilde and Touvron, Hugo and Misra, Ishan and J{\'e}gou, Herv{\'e} and Mairal, Julien and Bojanowski, Piotr and Joulin, Armand},
  booktitle={Proceedings of the IEEE/CVF international conference on computer vision},
  pages={9650--9660},
  year={2021}
}

@inproceedings{potje2024xfeat,
  title={{XFeat}: Accelerated features for lightweight image matching},
  author={Potje, Guilherme and Cadar, Felipe and Araujo, Andr{\'e} and Martins, Renato and Nascimento, Erickson R},
  booktitle={Proceedings of the IEEE/CVF Conference on Computer Vision and Pattern Recognition},
  pages={2682--2691},
  year={2024}
}

@article{wang2024journeybench,
  title={{JourneyBench}: a challenging one-stop vision-language understanding benchmark of generated images},
  author={Wang, Zhecan and Liu, Junzhang and Tang, Chia-Wei and Alomari, Hani and Sivakumar, Anushka and Sun, Rui and Li, Wenhao and Ayyubi, Hammad and You, Haoxuan and Ishmam, Alvi and others},
  journal={Advances in Neural Information Processing Systems},
  volume={37},
  pages={63110--63123},
  year={2024}
}

@article{jiang2024maven,
  title={{MaVEn}: An effective multi-granularity hybrid visual encoding framework for multimodal large language model},
  author={Jiang, Chaoya and Jia, Hongrui and Xu, Haiyang and Ye, Wei and Dong, Mengfan and Yan, Ming and Zhang, Ji and Huang, Fei and Zhang, Shikun},
  journal={Advances in Neural Information Processing Systems},
  volume={37},
  pages={101992--102010},
  year={2024}
}

@inproceedings{rasheed2024glamm,
  title={{GLaMM}: Pixel grounding large multimodal model},
  author={Rasheed, Hanoona and Maaz, Muhammad and Shaji, Sahal and Shaker, Abdelrahman and Khan, Salman and Cholakkal, Hisham and Anwer, Rao M and Xing, Eric and Yang, Ming-Hsuan and Khan, Fahad S},
  booktitle={Proceedings of the IEEE/CVF Conference on Computer Vision and Pattern Recognition},
  pages={13009--13018},
  year={2024}
}

@inproceedings{ma2024groma,
  title={Groma: Localized visual tokenization for grounding multimodal large language models},
  author={Ma, Chuofan and Jiang, Yi and Wu, Jiannan and Yuan, Zehuan and Qi, Xiaojuan},
  booktitle={European Conference on Computer Vision},
  pages={417--435},
  year={2024},
  organization={Springer}
}

@article{tong2024cambrian,
  title={Cambrian-1: A fully open, vision-centric exploration of multimodal {LLMs}},
  author={Tong, Shengbang and Brown, Ellis and Wu, Penghao and Woo, Sanghyun and Middepogu, Manoj and Akula, Sai C and Yang, Jihan and Yang, Shusheng and Iyer, Adithya and Pan, Xichen and others},
  journal={Advances in Neural Information Processing Systems},
  volume={37},
  pages={87310--87356},
  year={2024}
}

@inproceedings{wu2025f,
  title={{F-LMM}: Grounding frozen large multimodal models},
  author={Wu, Size and Jin, Sheng and Zhang, Wenwei and Xu, Lumin and Liu, Wentao and Li, Wei and Loy, Chen Change},
  booktitle={Proceedings of the IEEE/CVF Conference on Computer Vision and Pattern Recognition},
  pages={24710--24721},
  year={2025}
}

@inproceedings{yan2025task,
  title={Task preference optimization: Improving multimodal large language models with vision task alignment},
  author={Yan, Ziang and Li, Zhilin and He, Yinan and Wang, Chenting and Li, Kunchang and Li, Xinhao and Zeng, Xiangyu and Wang, Zilei and Wang, Yali and Qiao, Yu and others},
  booktitle={Proceedings of the Computer Vision and Pattern Recognition Conference},
  pages={29880--29892},
  year={2025}
}

@inproceedings{li2025llava,
  title={{LLaVA-ST}: A multimodal large language model for fine-grained spatial-temporal understanding},
  author={Li, Hongyu and Chen, Jinyu and Wei, Ziyu and Huang, Shaofei and Hui, Tianrui and Gao, Jialin and Wei, Xiaoming and Liu, Si},
  booktitle={Proceedings of the IEEE/CVF Conference on Computer Vision and Pattern Recognition},
  pages={8592--8603},
  year={2025}
}

@inproceedings{dong2025insight,
  title={{Insight-V}: Exploring long-chain visual reasoning with multimodal large language models},
  author={Dong, Yuhao and Liu, Zuyan and Sun, Hai-Long and Yang, Jingkang and Hu, Winston and Rao, Yongming and Liu, Ziwei},
  booktitle={Proceedings of the Computer Vision and Pattern Recognition Conference},
  pages={9062--9072},
  year={2025}
}

@inproceedings{gao2025mmat,
  title={{MMAT-1M}: A large reasoning dataset for multimodal agent tuning},
  author={Gao, Tianhong and Fu, Yannian and Wu, Weiqun and Yue, Haixiao and Liu, Shanshan and Zhang, Gang},
  booktitle={Proceedings of the IEEE/CVF International Conference on Computer Vision},
  pages={1484--1494},
  year={2025}
}

\appendix
\section*{Appendix}
\appendix
\section{Localization Quality Analysis}
\label{appendix:localization}

A natural concern is whether the MLLM, when acting as a spatial controller, produces region coordinates that actually cover the discriminative cues required for downstream verification. To assess localization quality, we let the base Qwen3-VL-4B model output region coordinates on all 12k samples in our evaluation set under both task formats. Under the pairwise task, the model produces a pair of bounding boxes, one per image. Under the multiple-choice task, the model first selects the most suspicious image among three references and then produces bounding boxes on both the selected image and the query image.

We then employ Qwen3-VL-30B-A3B-Instruct as an independent judge to evaluate whether each predicted bounding box covers the discriminative components relevant to instance identity and remains usable for subsequent local verification. The judge model is given the original image pair, the predicted bounding boxes, and the verification query, and is asked to determine whether the localization is acceptable. The judge accepts 94.22\% of the boxes in the pairwise task and 87.32\% in the multiple-choice task. The lower rate in the multiple-choice task reflects its additional step of selecting a candidate before localization.

\section{TWIN-Eval Regression Analysis}
\label{appendix:twin_regression}

We further analyze why LocalAgent falls below the base model on TWIN-Eval. Most of this gap already appears without any verifier. The no-tool CoT-RFT control, trained on the same data with accuracy and format rewards only, drops from 82.9\% to 76.5\% on this source, whereas LocalAgent reaches 74.5\%. To check whether the remaining change is uniform, we sample 1{,}000 TWIN-Eval examples, 500 pairwise and 500 multiple-choice, and use a larger MLLM as a blind stratification judge. Given the images, the question, and the ground-truth relation, the judge labels a sample as training-emphasis-aligned if it is a highly similar hard negative or a positive pair with substantial viewpoint, illumination, or background change, and assigns all remaining samples to the other group. The judge never sees predictions or correctness information from any evaluated model.

As shown in Table~\ref{tab:twin_stratified}, LocalAgent leads the base model by 8.8\% and CoT-RFT by 4.2\% on the training-emphasis-aligned cases, with a 43.8\% invocation rate. The aggregate regression concentrates in the other cases, where LocalAgent invokes LEV on 17.9\% of samples and reaches 69.2\% compared with 88.8\% for the base model and 78.2\% for CoT-RFT. LEV thus recovers performance on some difficult cases, but this descriptive analysis does not identify a single cause for the overall decrease. We therefore treat the benchmark-specific regression as a limitation of the current training and routing setup.

\begin{table}[h]
\centering
\caption{Stratified TWIN-Eval accuracy (\%) by visual characteristics associated with the training emphasis, with LocalAgent's LEV invocation rate per stratum.}
\label{tab:twin_stratified}
\small
\setlength{\tabcolsep}{8pt}
\begin{tabular}{lcccc}
\toprule
\textbf{TWIN-Eval subset} & \textbf{Base} & \textbf{CoT-RFT} & \textbf{LocalAgent} & \textbf{LEV invocation} \\
\midrule
Training-emphasis-aligned hard cases & 72.1 & 76.7 & \textbf{80.9} & 43.8 \\
Other cases              & \textbf{88.8} & 78.2 & 69.2 & 17.9 \\
\bottomrule
\end{tabular}
\end{table}

\section{Failure Mode Analysis}
\label{appendix:failure}

A qualitative audit of incorrect predictions identifies four recurring categories, which delimit where the framework does not yet work.

\begin{figure}[h]
    \centering
    \includegraphics[width=\linewidth]{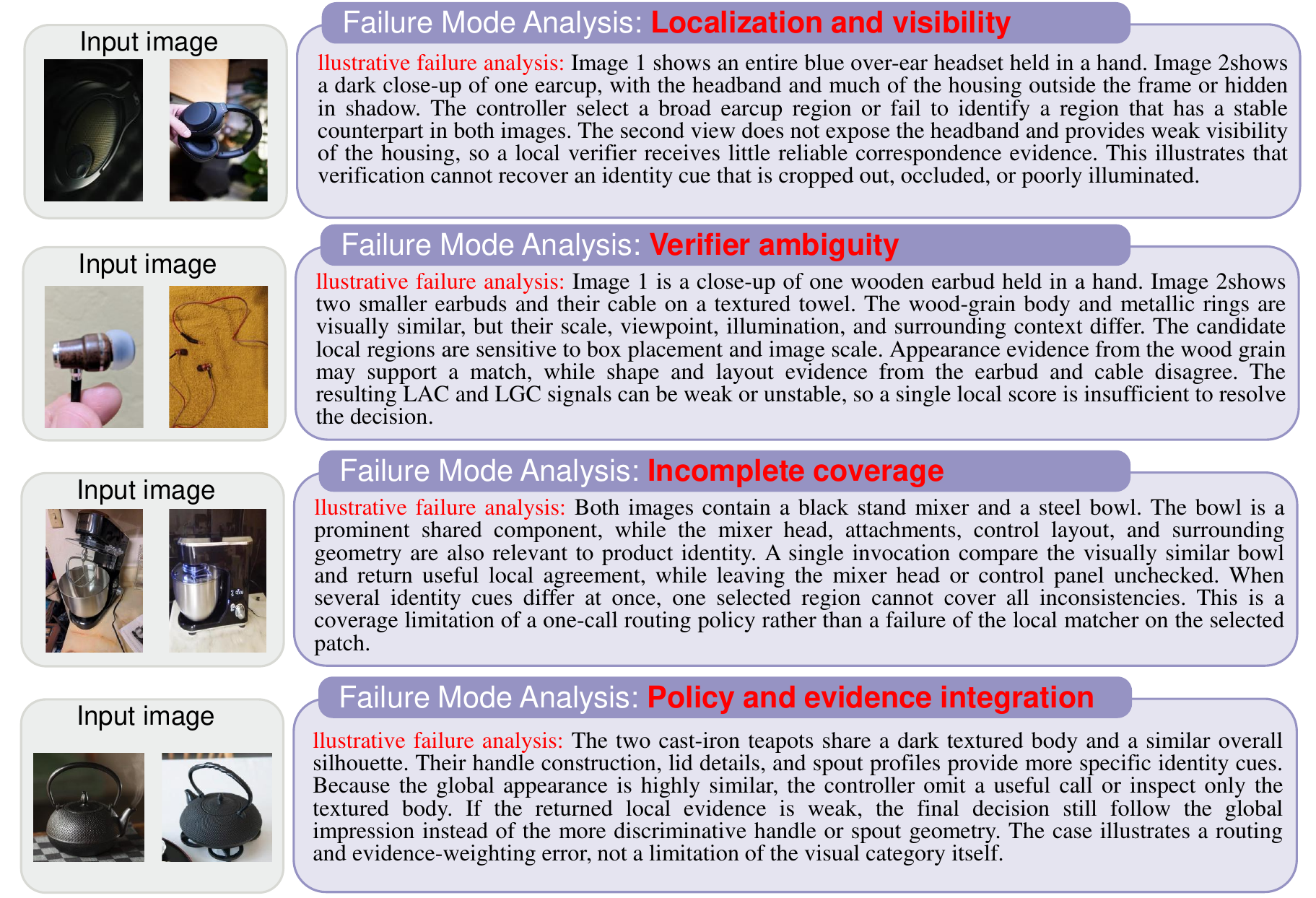}
    \caption{\textbf{Illustrative failure modes.}
 }
    \label{fig:failure_modes}
\end{figure}

\begin{itemize}
\item Localization and visibility. Predicted boxes miss the identifying region, or occlusion, blur, and extremely small details leave insufficient evidence for any verifier to act on.
\item Verifier ambiguity. Repetitive textures, large viewpoint changes, or sparse keypoints yield weak or mutually conflicting LAC and LGC evidence.
\item Incomplete coverage. A single invocation does not capture all inconsistent regions when several exist.
\item Policy and evidence integration. The MLLM omits a useful call, invokes an unnecessary verifier, or weighs the returned evidence incorrectly against its global impression.
\end{itemize}

These failure modes follow from the controller-bounded design and from the lack of joint evidence aggregation across several candidates in the multiple-choice task.

\begin{table}[h]
\centering
\caption{Sensitivity of LocalAgent to the auxiliary bounds on incorrect predictions.}
\label{tab:reward_sensitivity}
\small
\setlength{\tabcolsep}{8pt}
\begin{tabular}{lcc}
\toprule
\textbf{Setting} & $(\delta,\ \epsilon)$ & \textbf{Mean} \\
\midrule
\rowcolor{oursrow}
Default & (0.06, 0.12) & \textbf{80.7} \\
No auxiliary reward on incorrect answers & (0, 0) & 79.5 \\
Looser bounds & (0.10, 0.20) & 77.1 \\
\bottomrule
\end{tabular}
\end{table}

\section{Implementation Details}
\label{appendix:implementation}
We train the policy with GRPO based on Qwen3-VL-4B-Instruct using LoRA with rank 16 and $\alpha_{\text{LoRA}} = 32$ to the language backbone, while the visual encoder and the multimodal aligner are frozen. The main configuration uses 5 NVIDIA H20 GPUs, 1 for rollout and 4 for DDP training. Each prompt samples 8 rollout responses. We use AdamW with a learning rate of $1 \times 10^{-5}$, a weight decay of 0.1, and $\beta_1 = 0.9$, $\beta_2 = 0.95$. The learning rate follows cosine decay with a warmup ratio of 0.05, and gradient clipping is applied with a maximum norm of 0.5.

\section{Reward Sensitivity Analysis}
\label{appendix:reward_sensitivity}
We vary the auxiliary bounds $(\delta, \epsilon)$ in Eq.~\ref{eq:reward}, which apply to incorrect predictions, while keeping the reward coefficients fixed. Results are reported in Table~\ref{tab:reward_sensitivity}.  Removing the auxiliary reward on incorrect predictions lowers mean accuracy by 1.2\%, and loosening the bounds lowers it by 3.6\%. Bounding the auxiliary contribution on incorrect predictions therefore matters for the final performance.

\section{Statistical Reliability}
\label{appendix:statistical}

To assess the sample-level statistical reliability of the reported improvement, we align the LocalAgent and CoT-RFT predictions at the question level and compute paired-bootstrap 95\% confidence intervals and exact McNemar tests for the accuracy gain of LocalAgent over CoT-RFT. Results are reported in Table~\ref{tab:statistical}. The paired mean difference is positive and its confidence interval excludes zero in all three splits, and every McNemar test is significant, so the reported gain is not an artifact of a single evaluation run.

\begin{table}[h]
\centering
\caption{Paired-bootstrap 95\% confidence intervals and exact McNemar tests for the accuracy gain of LocalAgent over CoT-RFT.}
\label{tab:statistical}
\small
\setlength{\tabcolsep}{10pt}
\begin{tabular}{lcc}
\toprule
\textbf{Split} & \textbf{95\% bootstrap CI} & \textbf{McNemar $p$} \\
\midrule
Overall  & [4.27, 5.70] & $4.034\times 10^{-41}$ \\
Pairwise & [3.28, 5.07] & $7.849\times 10^{-20}$ \\
Multiple & [4.67, 6.95] & $5.181\times 10^{-23}$ \\
\bottomrule
\end{tabular}
\end{table}


\newpage
\section*{NeurIPS Paper Checklist}

\begin{enumerate}

\item {\bf Claims}
    \item[] Question: Do the main claims made in the abstract and introduction accurately reflect the paper's contributions and scope?
    \item[] Answer: \answerYes{} 
    \item[] Justification: The abstract and introduction state our three contributions (the collaborative agentic framework, the LEV toolkit with structured scalar evidence, and the auxiliary-capped RFT strategy), and the experimental results in Section~4 support the reported 4.9\% improvement in mean accuracy over the CoT-RFT baseline.
    \item[] Guidelines:
    \begin{itemize}
        \item The answer \answerNA{} means that the abstract and introduction do not include the claims made in the paper.
        \item The abstract and/or introduction should clearly state the claims made, including the contributions made in the paper and important assumptions and limitations. A \answerNo{} or \answerNA{} answer to this question will not be perceived well by the reviewers. 
        \item The claims made should match theoretical and experimental results, and reflect how much the results can be expected to generalize to other settings. 
        \item It is fine to include aspirational goals as motivation as long as it is clear that these goals are not attained by the paper. 
    \end{itemize}

\item {\bf Limitations}
    \item[] Question: Does the paper discuss the limitations of the work performed by the authors?
    \item[] Answer: \answerYes{} 
    \item[] Justification: We discuss limitations in Section~4, the Limitations paragraph of Section~5, and Appendices~\ref{appendix:twin_regression} and~\ref{appendix:failure}, including the regression on TWIN-Eval, the dependence on localization quality, and the lack of joint evidence aggregation in the multiple-choice task.
    \item[] Guidelines:
    \begin{itemize}
        \item The answer \answerNA{} means that the paper has no limitation while the answer \answerNo{} means that the paper has limitations, but those are not discussed in the paper. 
        \item The authors are encouraged to create a separate ``Limitations'' section in their paper.
        \item The paper should point out any strong assumptions and how robust the results are to violations of these assumptions (e.g., independence assumptions, noiseless settings, model well-specification, asymptotic approximations only holding locally). The authors should reflect on how these assumptions might be violated in practice and what the implications would be.
        \item The authors should reflect on the scope of the claims made, e.g., if the approach was only tested on a few datasets or with a few runs. In general, empirical results often depend on implicit assumptions, which should be articulated.
        \item The authors should reflect on the factors that influence the performance of the approach. For example, a facial recognition algorithm may perform poorly when image resolution is low or images are taken in low lighting. Or a speech-to-text system might not be used reliably to provide closed captions for online lectures because it fails to handle technical jargon.
        \item The authors should discuss the computational efficiency of the proposed algorithms and how they scale with dataset size.
        \item If applicable, the authors should discuss possible limitations of their approach to address problems of privacy and fairness.
        \item While the authors might fear that complete honesty about limitations might be used by reviewers as grounds for rejection, a worse outcome might be that reviewers discover limitations that aren't acknowledged in the paper. The authors should use their best judgment and recognize that individual actions in favor of transparency play an important role in developing norms that preserve the integrity of the community. Reviewers will be specifically instructed to not penalize honesty concerning limitations.
    \end{itemize}

\item {\bf Theory assumptions and proofs}
    \item[] Question: For each theoretical result, does the paper provide the full set of assumptions and a complete (and correct) proof?
    \item[] Answer: \answerNA{} 
    \item[] Justification: The paper presents an empirical framework and does not contain formal theoretical results requiring proofs.
    \item[] Guidelines:
    \begin{itemize}
        \item The answer \answerNA{} means that the paper does not include theoretical results. 
        \item All the theorems, formulas, and proofs in the paper should be numbered and cross-referenced.
        \item All assumptions should be clearly stated or referenced in the statement of any theorems.
        \item The proofs can either appear in the main paper or the supplemental material, but if they appear in the supplemental material, the authors are encouraged to provide a short proof sketch to provide intuition. 
        \item Inversely, any informal proof provided in the core of the paper should be complemented by formal proofs provided in appendix or supplemental material.
        \item Theorems and Lemmas that the proof relies upon should be properly referenced. 
    \end{itemize}

    \item {\bf Experimental result reproducibility}
    \item[] Question: Does the paper fully disclose all the information needed to reproduce the main experimental results of the paper to the extent that it affects the main claims and/or conclusions of the paper (regardless of whether the code and data are provided or not)?
    \item[] Answer: \answerYes{} 
    \item[] Justification: We specify the LocalAgent framework, the LEV toolkit (including the DINO and XFeat backbones), and the auxiliary-capped RFT objective with its reward components and bounds. Section~4 and Appendix~\ref{appendix:implementation} detail the training data, evaluation benchmarks, baselines, and hyperparameters.
    \item[] Guidelines:
    \begin{itemize}
        \item The answer \answerNA{} means that the paper does not include experiments.
        \item If the paper includes experiments, a \answerNo{} answer to this question will not be perceived well by the reviewers: Making the paper reproducible is important, regardless of whether the code and data are provided or not.
        \item If the contribution is a dataset and\slash or model, the authors should describe the steps taken to make their results reproducible or verifiable. 
        \item Depending on the contribution, reproducibility can be accomplished in various ways. For example, if the contribution is a novel architecture, describing the architecture fully might suffice, or if the contribution is a specific model and empirical evaluation, it may be necessary to either make it possible for others to replicate the model with the same dataset, or provide access to the model. In general. releasing code and data is often one good way to accomplish this, but reproducibility can also be provided via detailed instructions for how to replicate the results, access to a hosted model (e.g., in the case of a large language model), releasing of a model checkpoint, or other means that are appropriate to the research performed.
        \item While NeurIPS does not require releasing code, the conference does require all submissions to provide some reasonable avenue for reproducibility, which may depend on the nature of the contribution. For example
        \begin{enumerate}
            \item If the contribution is primarily a new algorithm, the paper should make it clear how to reproduce that algorithm.
            \item If the contribution is primarily a new model architecture, the paper should describe the architecture clearly and fully.
            \item If the contribution is a new model (e.g., a large language model), then there should either be a way to access this model for reproducing the results or a way to reproduce the model (e.g., with an open-source dataset or instructions for how to construct the dataset).
            \item We recognize that reproducibility may be tricky in some cases, in which case authors are welcome to describe the particular way they provide for reproducibility. In the case of closed-source models, it may be that access to the model is limited in some way (e.g., to registered users), but it should be possible for other researchers to have some path to reproducing or verifying the results.
        \end{enumerate}
    \end{itemize}

\item {\bf Open access to data and code}
    \item[] Question: Does the paper provide open access to the data and code, with sufficient instructions to faithfully reproduce the main experimental results, as described in supplemental material?
    \item[] Answer: \answerYes{} 
    \item[] Justification: The codes are publicly available at \url{https://github.com/Firefly-zoo/LocalAgent-main}. All evaluation data come from public benchmarks.
    \item[] Guidelines:
    \begin{itemize}
        \item The answer \answerNA{} means that paper does not include experiments requiring code.
        \item Please see the NeurIPS code and data submission guidelines (\url{https://neurips.cc/public/guides/CodeSubmissionPolicy}) for more details.
        \item While we encourage the release of code and data, we understand that this might not be possible, so \answerNo{} is an acceptable answer. Papers cannot be rejected simply for not including code, unless this is central to the contribution (e.g., for a new open-source benchmark).
        \item The instructions should contain the exact command and environment needed to run to reproduce the results. See the NeurIPS code and data submission guidelines (\url{https://neurips.cc/public/guides/CodeSubmissionPolicy}) for more details.
        \item The authors should provide instructions on data access and preparation, including how to access the raw data, preprocessed data, intermediate data, and generated data, etc.
        \item The authors should provide scripts to reproduce all experimental results for the new proposed method and baselines. If only a subset of experiments are reproducible, they should state which ones are omitted from the script and why.
        \item At submission time, to preserve anonymity, the authors should release anonymized versions (if applicable).
        \item Providing as much information as possible in supplemental material (appended to the paper) is recommended, but including URLs to data and code is permitted.
    \end{itemize}

\item {\bf Experimental setting/details}
    \item[] Question: Does the paper specify all the training and test details (e.g., data splits, hyperparameters, how they were chosen, type of optimizer) necessary to understand the results?
    \item[] Answer: \answerYes{} 
    \item[] Justification: Section~4.1 and Appendix~\ref{appendix:implementation} describe the data, protocols, and training hyperparameters, and Appendix~\ref{appendix:reward_sensitivity} reports the sensitivity to the auxiliary bounds.
    \item[] Guidelines:
    \begin{itemize}
        \item The answer \answerNA{} means that the paper does not include experiments.
        \item The experimental setting should be presented in the core of the paper to a level of detail that is necessary to appreciate the results and make sense of them.
        \item The full details can be provided either with the code, in appendix, or as supplemental material.
    \end{itemize}

\item {\bf Experiment statistical significance}
    \item[] Question: Does the paper report error bars suitably and correctly defined or other appropriate information about the statistical significance of the experiments?
    \item[] Answer: \answerYes{} 
    \item[] Justification: Appendix~\ref{appendix:statistical} reports paired-bootstrap 95\% confidence intervals and exact McNemar tests for LocalAgent's gain over CoT-RFT, which are significant on the overall, pairwise, and multiple-choice splits.
    \item[] Guidelines:
    \begin{itemize}
        \item The answer \answerNA{} means that the paper does not include experiments.
        \item The authors should answer \answerYes{} if the results are accompanied by error bars, confidence intervals, or statistical significance tests, at least for the experiments that support the main claims of the paper.
        \item The factors of variability that the error bars are capturing should be clearly stated (for example, train/test split, initialization, random drawing of some parameter, or overall run with given experimental conditions).
        \item The method for calculating the error bars should be explained (closed form formula, call to a library function, bootstrap, etc.)
        \item The assumptions made should be given (e.g., Normally distributed errors).
        \item It should be clear whether the error bar is the standard deviation or the standard error of the mean.
        \item It is OK to report 1-sigma error bars, but one should state it. The authors should preferably report a 2-sigma error bar than state that they have a 96\% CI, if the hypothesis of Normality of errors is not verified.
        \item For asymmetric distributions, the authors should be careful not to show in tables or figures symmetric error bars that would yield results that are out of range (e.g., negative error rates).
        \item If error bars are reported in tables or plots, the authors should explain in the text how they were calculated and reference the corresponding figures or tables in the text.
    \end{itemize}

\item {\bf Experiments compute resources}
    \item[] Question: For each experiment, does the paper provide sufficient information on the computer resources (type of compute workers, memory, time of execution) needed to reproduce the experiments?
    \item[] Answer: \answerYes{} 
    \item[] Justification: Experiment compute resources are reported in Appendix~\ref{appendix:implementation}.
    \item[] Guidelines:
    \begin{itemize}
        \item The answer \answerNA{} means that the paper does not include experiments.
        \item The paper should indicate the type of compute workers CPU or GPU, internal cluster, or cloud provider, including relevant memory and storage.
        \item The paper should provide the amount of compute required for each of the individual experimental runs as well as estimate the total compute. 
        \item The paper should disclose whether the full research project required more compute than the experiments reported in the paper (e.g., preliminary or failed experiments that didn't make it into the paper). 
    \end{itemize}
    
\item {\bf Code of ethics}
    \item[] Question: Does the research conducted in the paper conform, in every respect, with the NeurIPS Code of Ethics \url{https://neurips.cc/public/EthicsGuidelines}?
    \item[] Answer: \answerYes{} 
    \item[] Justification: The research uses publicly available models and benchmarks and conforms with the NeurIPS Code of Ethics.
    \item[] Guidelines:
    \begin{itemize}
        \item The answer \answerNA{} means that the authors have not reviewed the NeurIPS Code of Ethics.
        \item If the authors answer \answerNo, they should explain the special circumstances that require a deviation from the Code of Ethics.
        \item The authors should make sure to preserve anonymity (e.g., if there is a special consideration due to laws or regulations in their jurisdiction).
    \end{itemize}

\item {\bf Broader impacts}
    \item[] Question: Does the paper discuss both potential positive societal impacts and negative societal impacts of the work performed?
    \item[] Answer: \answerYes{} 
    \item[] Justification: LocalAgent narrows the performance gap between open-source and closed-source MLLMs on fine-grained instance-level verification, contributing to the accessibility of capable visual reasoning systems. The work poses no foreseeable negative societal impact beyond standard considerations applicable to general-purpose MLLMs.
    \item[] Guidelines:
    \begin{itemize}
        \item The answer \answerNA{} means that there is no societal impact of the work performed.
        \item If the authors answer \answerNA{} or \answerNo, they should explain why their work has no societal impact or why the paper does not address societal impact.
        \item Examples of negative societal impacts include potential malicious or unintended uses (e.g., disinformation, generating fake profiles, surveillance), fairness considerations (e.g., deployment of technologies that could make decisions that unfairly impact specific groups), privacy considerations, and security considerations.
        \item The conference expects that many papers will be foundational research and not tied to particular applications, let alone deployments. However, if there is a direct path to any negative applications, the authors should point it out. For example, it is legitimate to point out that an improvement in the quality of generative models could be used to generate Deepfakes for disinformation. On the other hand, it is not needed to point out that a generic algorithm for optimizing neural networks could enable people to train models that generate Deepfakes faster.
        \item The authors should consider possible harms that could arise when the technology is being used as intended and functioning correctly, harms that could arise when the technology is being used as intended but gives incorrect results, and harms following from (intentional or unintentional) misuse of the technology.
        \item If there are negative societal impacts, the authors could also discuss possible mitigation strategies (e.g., gated release of models, providing defenses in addition to attacks, mechanisms for monitoring misuse, mechanisms to monitor how a system learns from feedback over time, improving the efficiency and accessibility of ML).
    \end{itemize}
    
\item {\bf Safeguards}
    \item[] Question: Does the paper describe safeguards that have been put in place for responsible release of data or models that have a high risk for misuse (e.g., pre-trained language models, image generators, or scraped datasets)?
    \item[] Answer: \answerNA{} 
    \item[] Justification: The paper does not release models or datasets with a high risk of misuse. The framework is built on publicly available models and standard benchmarks.
    \item[] Guidelines:
    \begin{itemize}
        \item The answer \answerNA{} means that the paper poses no such risks.
        \item Released models that have a high risk for misuse or dual-use should be released with necessary safeguards to allow for controlled use of the model, for example by requiring that users adhere to usage guidelines or restrictions to access the model or implementing safety filters. 
        \item Datasets that have been scraped from the Internet could pose safety risks. The authors should describe how they avoided releasing unsafe images.
        \item We recognize that providing effective safeguards is challenging, and many papers do not require this, but we encourage authors to take this into account and make a best faith effort.
    \end{itemize}

\item {\bf Licenses for existing assets}
    \item[] Question: Are the creators or original owners of assets (e.g., code, data, models), used in the paper, properly credited and are the license and terms of use explicitly mentioned and properly respected?
    \item[] Answer: \answerYes{} 
    \item[] Justification: All external assets used in this work are properly cited in the references with their original publications.
    \item[] Guidelines:
    \begin{itemize}
        \item The answer \answerNA{} means that the paper does not use existing assets.
        \item The authors should cite the original paper that produced the code package or dataset.
        \item The authors should state which version of the asset is used and, if possible, include a URL.
        \item The name of the license (e.g., CC-BY 4.0) should be included for each asset.
        \item For scraped data from a particular source (e.g., website), the copyright and terms of service of that source should be provided.
        \item If assets are released, the license, copyright information, and terms of use in the package should be provided. For popular datasets, \url{paperswithcode.com/datasets} has curated licenses for some datasets. Their licensing guide can help determine the license of a dataset.
        \item For existing datasets that are re-packaged, both the original license and the license of the derived asset (if it has changed) should be provided.
        \item If this information is not available online, the authors are encouraged to reach out to the asset's creators.
    \end{itemize}

\item {\bf New assets}
    \item[] Question: Are new assets introduced in the paper well documented and is the documentation provided alongside the assets?
    \item[] Answer: \answerNA{} 
    \item[] Justification: This paper does not release new public assets. All datasets used are pre-existing public benchmarks and properly cited.
    \item[] Guidelines:
    \begin{itemize}
        \item The answer \answerNA{} means that the paper does not release new assets.
        \item Researchers should communicate the details of the dataset\slash code\slash model as part of their submissions via structured templates. This includes details about training, license, limitations, etc. 
        \item The paper should discuss whether and how consent was obtained from people whose asset is used.
        \item At submission time, remember to anonymize your assets (if applicable). You can either create an anonymized URL or include an anonymized zip file.
    \end{itemize}

\item {\bf Crowdsourcing and research with human subjects}
    \item[] Question: For crowdsourcing experiments and research with human subjects, does the paper include the full text of instructions given to participants and screenshots, if applicable, as well as details about compensation (if any)? 
    \item[] Answer: \answerNA{} 
    \item[] Justification: The work does not involve crowdsourcing or research with human subjects.
    \item[] Guidelines:
    \begin{itemize}
        \item The answer \answerNA{} means that the paper does not involve crowdsourcing nor research with human subjects.
        \item Including this information in the supplemental material is fine, but if the main contribution of the paper involves human subjects, then as much detail as possible should be included in the main paper. 
        \item According to the NeurIPS Code of Ethics, workers involved in data collection, curation, or other labor should be paid at least the minimum wage in the country of the data collector. 
    \end{itemize}

\item {\bf Institutional review board (IRB) approvals or equivalent for research with human subjects}
    \item[] Question: Does the paper describe potential risks incurred by study participants, whether such risks were disclosed to the subjects, and whether Institutional Review Board (IRB) approvals (or an equivalent approval/review based on the requirements of your country or institution) were obtained?
    \item[] Answer: \answerNA{} 
    \item[] Justification: The work does not involve human subjects and therefore does not require IRB review.
    \item[] Guidelines:
    \begin{itemize}
        \item The answer \answerNA{} means that the paper does not involve crowdsourcing nor research with human subjects.
        \item Depending on the country in which research is conducted, IRB approval (or equivalent) may be required for any human subjects research. If you obtained IRB approval, you should clearly state this in the paper. 
        \item We recognize that the procedures for this may vary significantly between institutions and locations, and we expect authors to adhere to the NeurIPS Code of Ethics and the guidelines for their institution. 
        \item For initial submissions, do not include any information that would break anonymity (if applicable), such as the institution conducting the review.
    \end{itemize}

\item {\bf Declaration of LLM usage}
    \item[] Question: Does the paper describe the usage of LLMs if it is an important, original, or non-standard component of the core methods in this research? Note that if the LLM is used only for writing, editing, or formatting purposes and does \emph{not} impact the core methodology, scientific rigor, or originality of the research, declaration is not required.
    \item[] Answer: \answerNA{} 
    \item[] Justification: The core method does not involve LLMs as important, original, or non-standard components. LLMs were used only for writing, editing, or formatting.
    \item[] Guidelines:
    \begin{itemize}
        \item The answer \answerNA{} means that the core method development in this research does not involve LLMs as any important, original, or non-standard components.
        \item Please refer to our LLM policy in the NeurIPS handbook for what should or should not be described.
    \end{itemize}

\end{enumerate}

\end{document}